\documentclass[conference]{IEEEtran}
\IEEEoverridecommandlockouts
\usepackage{cite}
\usepackage{amsmath,amssymb,amsfonts}
\usepackage{algorithmic}
\usepackage{graphicx}
\usepackage[lofdepth,lotdepth]{subfig}
\usepackage[font=footnotesize,labelfont=bf]{caption}
\usepackage{textcomp}
\usepackage{xcolor}
\usepackage{float}
\usepackage{enumitem}
\usepackage{booktabs}
\usepackage{multirow}
\usepackage[square,numbers]{natbib}
\usepackage{hyperref}
\usepackage{cleveref}
\usepackage{array}
\newcolumntype{?}{!{\vrule width 1pt}}

\usepackage{pifont}                     
\newcommand{\xmark}{\ding{55}}%

\def\BibTeX{{\rm B\kern-.05em{\sc i\kern-.025em b}\kern-.08em
    T\kern-.1667em\lower.7ex\hbox{E}\kern-.125emX}}

\usepackage{fancyhdr,lipsum}
\fancypagestyle{firstpage}{
  \fancyhf{}
  \fancyhead[C]{Published at the 2020 International Joint Conference on Neural Networks (IJCNN), Glasgow, Scotland, July 2020.}
  \fancyfoot[C]{\thepage}
}

\begin{document}

\title{An Efficient Spiking Neural Network for Recognizing Gestures with a DVS Camera\\
on the Loihi Neuromorphic Processor\\
\vspace*{-10pt}
}


\author{\IEEEauthorblockN{Riccardo Massa$^{1,2,*}$\thanks{*These authors contributed equally to this work.}, Alberto Marchisio$^{1,*}$, Maurizio Martina$^2$, Muhammad Shafique$^1$}
\IEEEauthorblockA{\textit{$^1$Technische Universit{\"a}t Wien, Vienna, Austria}\ \ \ \ \ \ \ \textit{$^2$Politecnico di Torino, Turin, Italy}} 
\IEEEauthorblockA{\textit{Email: s251880@studenti.polito.it, \{alberto.marchisio, muhammad.shafique\}@tuwien.ac.at, maurizio.martina@polito.it}\\
}\vspace*{-30pt}}

\maketitle
\thispagestyle{firstpage}

\begin{small}
\begin{abstract}
Spiking Neural Networks (SNNs), the third generation NNs, have come under the spotlight for machine learning based applications due to their biological plausibility and reduced complexity compared to traditional artificial Deep Neural Networks (DNNs). These SNNs can be implemented with extreme energy efficiency on neuromorphic processors like the Intel Loihi research chip, and fed by event-based sensors, such as DVS cameras. However, DNNs with many layers can achieve relatively high accuracy on image classification and recognition tasks, as the research on learning rules for SNNs for real-world applications is still not mature. The accuracy results for SNNs are typically obtained either by converting the trained DNNs into SNNs, or by directly designing and training SNNs in the spiking domain.
Towards the conversion from a DNN to an SNN, we perform a comprehensive analysis of such process, specifically designed for Intel Loihi, showing our methodology for the design of an SNN that achieves nearly the same accuracy results as its corresponding DNN. Towards the usage of the event-based sensors, we design a pre-processing method, evaluated for the DvsGesture dataset, which makes it possible to be used in the DNN domain. Hence, based on the outcome of the first analysis, we train a DNN for the pre-processed DvsGesture dataset, and convert it into the spike domain for its deployment on Intel Loihi, which enables real-time gesture recognition. The results show that our SNN achieves 89.64\% classification accuracy and occupies only 37 Loihi cores. The source code for generating our experiments is available \href{https://github.com/albertomarchisio/EfficientSNN}{online}\footnote{\url{https://github.com/albertomarchisio/EfficientSNN}} for reproducible research.

\end{abstract}

\end{small}

\vspace*{-5pt}
\section{Introduction}
Recent developments of artificial Deep Neural Networks (DNNs) have pushed forward the state-of-the-art in the field of image recognition~\cite{resnet}. However, the high power demand required by these networks when it comes to perform inference tasks on the edge devices~\cite{Marchisio2019DL4EC}\cite{Shafique2020RobustML} limits the spread of DNNs in scenarios/use-cases where the energy/power consumption is crucial~\cite{Shafique2018NextGenML}\cite{Zhang2019RobustML}.
On the other hand, Spiking Neural Networks (SNNs), due to their biologically plausible model, have shown promising results both in terms of power/energy efficiency and real-time classification performance~\cite{Zambrano}. By leveraging the spike-based communication between neurons, SNNs guarantee a lower computational load, as well as a reduction in the latency. As a side effect, SNNs have also shown a different behavior than DNNs when threatened by adversarial attacks~\cite{Marchisio2019SNNUnderAttack}.
\\
Along with the development of efficient SNN specialized accelerators (like TrueNorth~\cite{truenorth}, SpiNNaker~\cite{spinnaker} and Intel Loihi~\cite{loihi}), another advancement in the field of neuromorphic hardware has come from a new generation of camera, the DVS event-based sensor~\cite{DVS}. Such a device, differently from a classical frame-based camera, works emulating the behavior of the human retina. Thus, the recorded information is not a series of time-wise separated frames, but a sequence of spikes, which are generated every time a change of light intensity is detected. 
The event-based behavior of these sensors pairs well with SNNs, i.e., the output of a DVS camera can be used as the input of an SNN to elaborate events in real-time.

A promising approach to train SNNs in a supervised learning scenario is to train a DNN with state-of-the-art backpropagation approaches, and then assign the trained parameters (weights and biases) to an equivalent SNN representation by applying a conversion process.
This approach has shown promising results~\cite{rueckauer}, mostly because it allows to get the best from the two worlds: the converted SNN totally behaves like a normal SNN, with its benefits in terms of efficiency and latency. At the same time, the network has been trained with efficient methodologies that ensure good results in classification tasks.
However, such a conversion may not always provide the expected results. In fact, many aspects have to be taken into account, like the original DNN structure, the training process, as well as the parameters that control the DNN-to-SNN conversion.
This is especially true when the converted SNN has to be deployed on a limited precision hardware like Intel Loihi, which restricts the degree of freedom of the conversion process.

\begin{figure}[!h]
    \centering
    \includegraphics[scale=1.1]{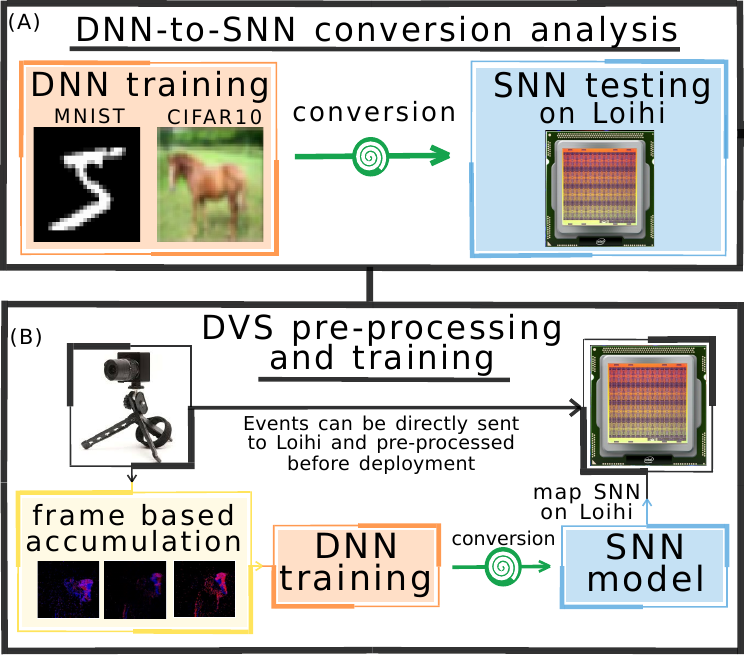}
    \caption{Workflow of our research.}
    \label{introduction_image}
    \vspace*{-15pt}
\end{figure}

Towards this, in this paper, we present a complete DNN-to-SNN design process (Figure~\ref{introduction_image}A), systematically discussing the effects of the key parameters that are used in the conversion. We evaluate their effect, and extract important general rules that can be successfully applied when it comes to develop an SNN for Intel Loihi or similar neuromorphic processors.
Once we have an SNN that gets good accuracy results both on the MNIST~\cite{MNIST} and the CIFAR10~\cite{CIFAR10} datasets, we evaluate it also on the DvsGesture dataset~\cite{DVSgesture}, which comprise 11 gestures recorded with a DVS event-based camera (Figure~\ref{introduction_image}B). 
The main challenge when adopting the DNN-to-SNN conversion approach to get a trained SNN is that we cannot train a DNN on the event-series coming from the DVS camera. For this reason, we first need to collect the events into frames, and then train the DNN on such converted dataset.
Different pre-processing techniques are discussed in this paper, also reporting the accuracy results achieved by the DNN on the generated converted dataset. Finally, after performing the conversion, the SNN is tested on the DvsGesture dataset, and afterwards, it is ready to be deployed for real-time classification on Intel Loihi.

\textbf{In a nutshell, our Novel Contributions are:}
\vspace*{-2pt}
\begin{itemize}[leftmargin=*]
    \item We perform a comprehensive parameter analysis of the process of converting a DNN into an SNN. (\textbf{Section \ref{ANN SNN conversion analysis}})
    \item We design a pre-processing method for the DvsGesture dataset through frame-based accumulation, to make such dataset compatible with the DNN domain. (\textbf{Section~\ref{dvs_gesture_pre-processing}})
    \item We train a given DNN for the pre-processed DvsGesture dataset and convert it to an SNN that can then be deployed on Intel Loihi. (\textbf{Section \ref{results DVS}})
\end{itemize}

Before proceeding to the technical sections, in \textbf{Section~\ref{sec:background}} we present an overview of the SNNs, the Intel Loihi research chip, and of the DNN-to-SNN conversion approach, to a level of detail necessary to understand the contributions of this paper.

\section{Background and Related Work}
\label{sec:background}

\subsection{Spiking Neural Networks}
\begin{figure*}[!h]
    \centering
    \includegraphics[scale=0.68]{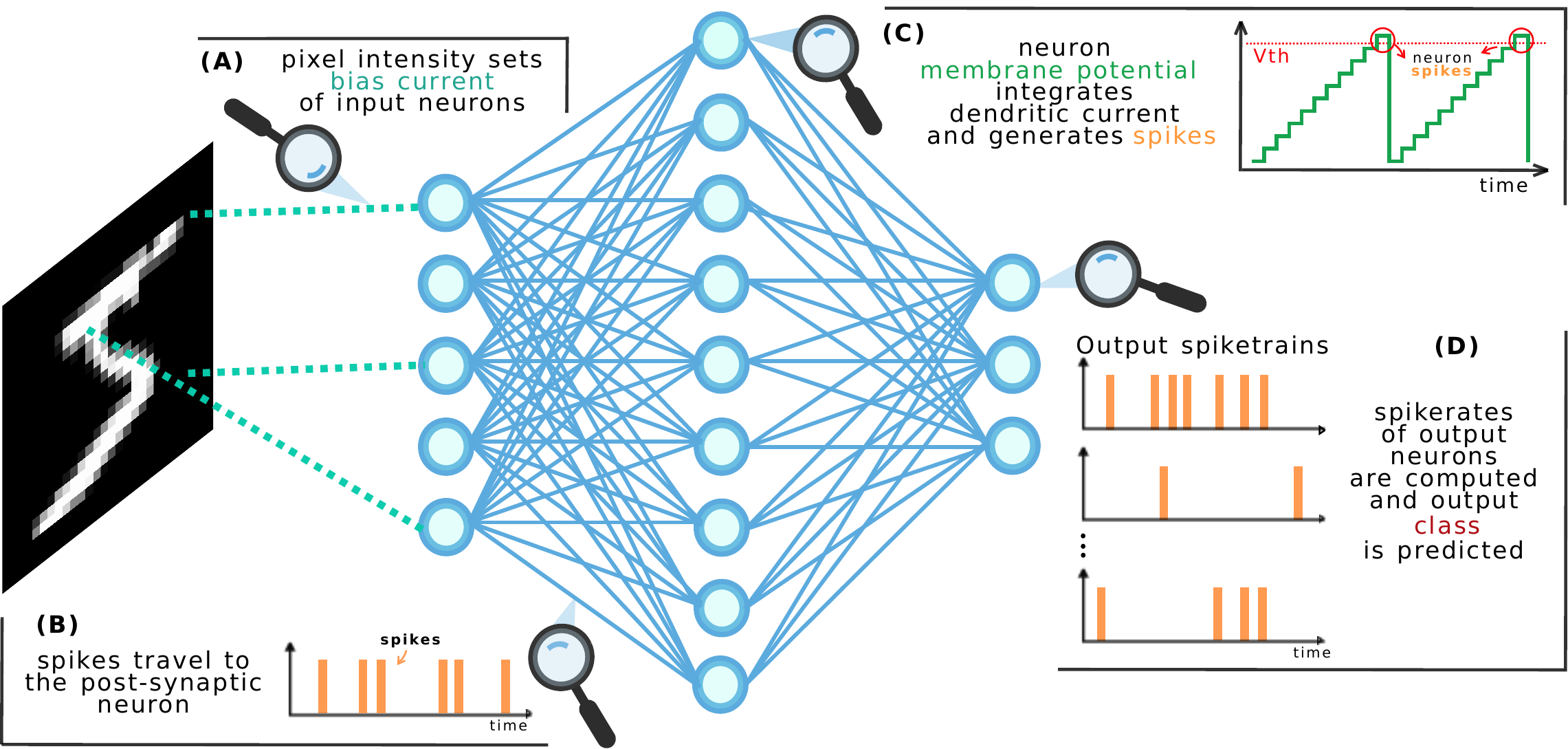}
    \caption{Structure and main steps for an SNN for image classification task. In this example, a fully-connected SNN, with neurons represented as circles and synapses as lines, is shown. \textbf{(A)} The input image is not converted with a spike-encoding algorithm, but pixel intensities set the bias currents of neurons in the input layer. \textbf{(B)} The spikes from the pre-synaptic neuron travel across the synapse and accumulate in the dendritic tree of post-synaptic neurons. The membrane current of the post-synaptic neuron integrates the incoming weighted spike trains. \textbf{(C)} The neuron membrane potential integrates the bias current and the membrane current. An output spike is generated each time the potential reaches a predefined threshold. Afterwards, the membrane potential is set back to the initial level. \textbf{(D)} The output neurons, one for each class, generate spike trains. For each neuron, its corresponding spikerate in a predefined time-window is computed, which is then used as the output prediction for its class.}
    \label{SNN_scheme}
    \vspace*{-15pt}
\end{figure*}

Spiking Neural Networks (SNNs) are based on the biologically plausible models of neurons~\cite{Pfeiffer}, which communicate asynchronously through series of spikes.
The structure and behavior of an SNN are presented in Figure~\ref{SNN_scheme}.

The SNNs' major improvements over traditional DNNs are the following~\cite{Pfeiffer}:
\begin{itemize}[leftmargin=*]
    \item The intrinsic asynchronous, spike-based communication protocol adopted in the network allows to \textit{reduce the power/energy consumption} required in the computations and communication. 
    \item The asynchronous, spike-based design makes this networks ideal to \textit{cooperate with event-based sensors}. The events provided as an input can be seen as a train of spikes directly processable by the network.  
\end{itemize}

The SNNs' main weakness relies on the fact that the classical supervised learning approach, i.e., the backpropagation, cannot be applied due to the non-differentiability of the SNN loss function ~\cite{ClosingTheAccuracyGap}.
Therefore, two main approaches have been proposed to achieve supervised learning in SNNs:
\begin{itemize}[leftmargin=*]
    \item Use the backpropagation algorithm directly in the spiking domain. This method generally requires to substitute the loss function with a placeholder function, that can be differentiated~\cite{SLAYERtheory}\cite{SLAYER}.
    \item Train an equivalent DNN model and then convert it to an SNN in the spiking domain.
\end{itemize}

In this article, we focus on the latter approach. Training the network in the DNN domain allows us to use the current state-of-the-art training policies and techniques. 
Moreover, the DNN-to-SNN conversion technique has shown promising results, allowing to get SNNs that reach the same, or very close levels of accuracy, compared to their corresponding DNN versions~\cite{ClosingTheAccuracyGap}\cite{rueckauer}.
However, some precautions and limitations have to be considered when using this approach, as we will explain in our analysis in Section~\ref{ANN SNN conversion analysis}. 
\subsection{Intel Loihi Neuromorphic Research Chip}
DNNs achieve the best results in terms of accuracy and efficiency when executed on highly parallel hardware like GPUs, and even more with specialized hardware accelerators, like Google TPU~\cite{TPU}, MPNA~\cite{MPNA}. Similarly, SNNs require their specialized hardware to achieve the best results in terms of power/energy efficiency and latency~\cite{SNNhardware}. 
Neuromorphic chips represent an efficient hardware solution when it comes to the implementation of SNNs. Unlike the artificial neuron model and synchronous structure of traditional DNNs, the highly parallel asynchronous structure, combined with the hardware implementation of a biologically plausible neuron model, such as the \textit{leaky-integrate-and-fire (LIF)} model~\cite{STDP}, allows to achieve far better results both in latency and power/energy efficiency with SNNs when compared to their CPU and GPU implementation.  
Recent developments in the field of neuromorphic hardware have brought valid and powerful solutions for the deployment of SNN models, like IBM TrueNorth~\cite{truenorth}, SpiNNaker~\cite{spinnaker} and Intel Loihi~\cite{loihi}. 

In this paper, we focus on the \textit{Intel Loihi}~\cite{loihi}, which is a neuromorphic processor providing highly parallel and energy efficient asynchronous computation. The chip comprise a neuromorphic mesh of 128 neurocores, and 3 x86 processors, as well as an asynchronous network-on-chip (NoC) that connects neurocores allowing neuron-to-neuron communication.
Each neurocore implements up to 1024 spiking neural compartments units, such that the compartments can be combined in a tree structure to form multi-compartment neurons. Neurons variables are updated at every algorithmic time-step. The spikes generated by a neuron are delivered to all the compartments belonging to its synaptic fan-out through the NoC. The NoC allows to deliver spikes between different neurocores in a packet-messaged form, following a mesh operation that is executed over a series of algorithmic time-steps. In the absence of a global clock, a barrier synchronization mechanism is used to ensure that at the end of each time-step all neurons are synchronized.  
An off-chip communication interface allows to extend the mesh up to 4096 on-chip cores, and up to 16,384 hierarchically connected cores.

\begin{figure}[!h]
    \centering
    \includegraphics[scale=0.09]{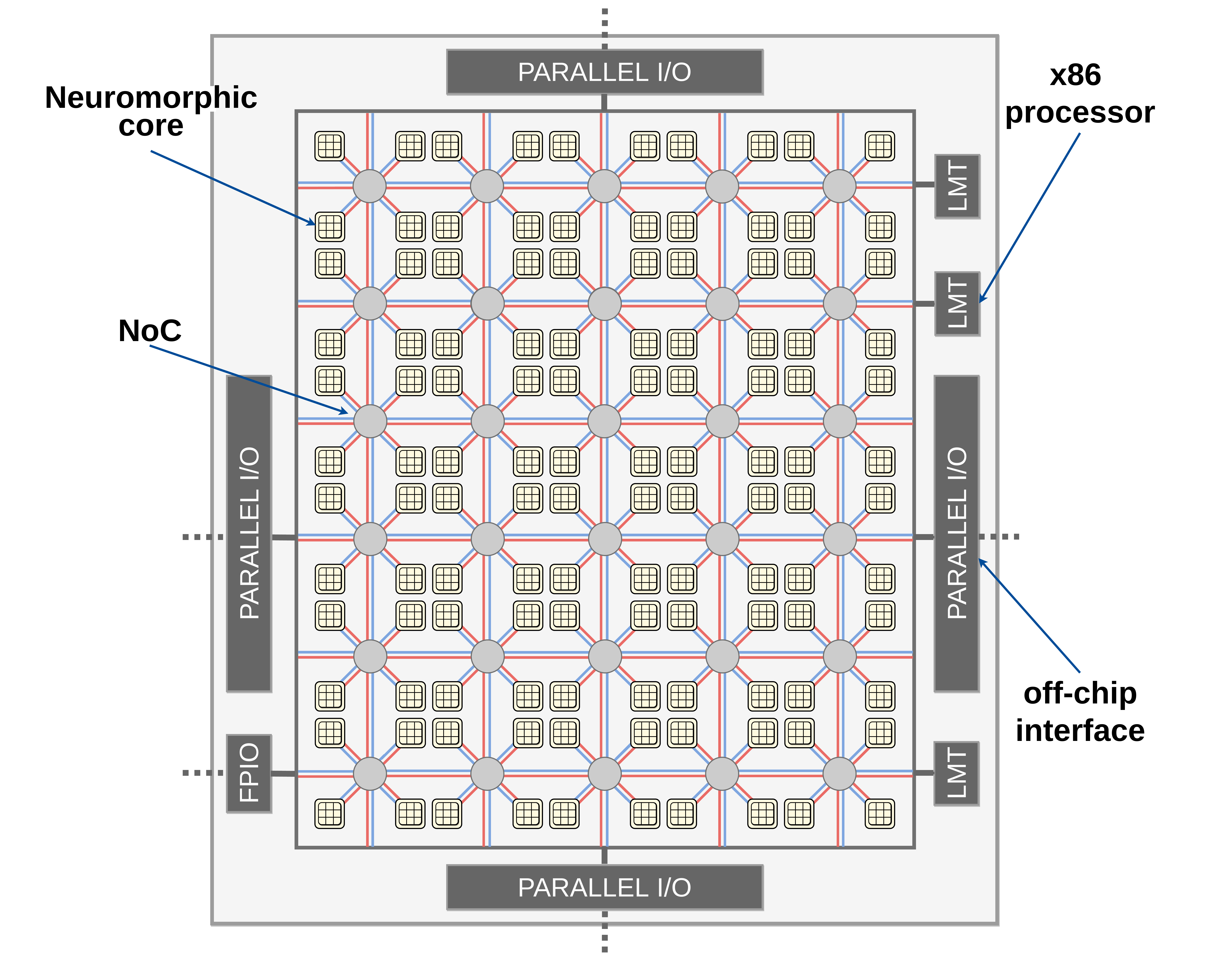}
    \caption{Loihi single chip architecture~\cite{loihi}.}
    \label{LOIHI_CHIP}
    \vspace*{-15pt}
\end{figure}

The architecture of a single Loihi chip is displayed in Figure~\ref{LOIHI_CHIP}. The biologically-plausible neuron model adopted by the Loihi architecture is based on a modified version of the CUBA leaky-integrate-and-fire model~\cite{loihi}.
More specifically, each neuron is represented as a dendritic compartment, which receives the incoming spikes from the pre-synaptic neurons.


Each neuron is characterized by its compartment current $u_{c}(t)$ and its compartment membrane potential $v_c(t)$~\cite{loihi}.
Given a postsynaptic neuron $n_i$, it receives an input train of spikes from presynaptic neuron $n_j$ that can be represented as a train of Dirac delta functions: 
$\sigma_j(t) = \sum_{k}{\delta(t-t_k)}$ 
where $t_k$ is the spike time.

The train of spikes is first filtered by a synaptic filter input response $\alpha_u(t)$ and then multiplied by the synaptic weight $w_{ij}$ associated to the synapse that connects neurons $n_i$ and $n_j$. The synaptic response current can then be computed as the sum of all the weighted and filtered spike trains, with an additional bias current $u_{bias_i}$:
\vspace*{-5pt}
$$
u_{c_i}(t) = \sum_{j}{w_{ij}(\alpha_u*\sigma_j)(t)} + u_{bias_i}
\vspace*{-5pt}
$$

Finally, the synaptic current is integrated by the membrane potential $v_{c_i}(t)$.
\vspace*{-5pt}
$$
\dot{v_{c_i}}(t) = -\frac{1}{\tau_v}v_{c_i}(t) + u_{c_i}(t) -V_{th_i}\sigma_i(t) 
\vspace*{-5pt}
$$

When the membrane potential reaches a threshold $V_{th}$, the neuron spikes. Then, the membrane potential is reset to a $V_{rest}$ value and starts increasing again as new input spikes are received.
The time constant $\tau_v$ is responsible for the \textit{leaky} behavior of the model~\cite{loihi}.

\subsection{DNN-to-SNN Conversion}
\label{section ANN_SNN_conversion}
The DNN-to-SNN conversion approach has shown promising results in terms of accuracy consistency among the original DNN and the converted SNN~\cite{rueckauer}. To reach such results, the trained parameters of the DNN must be efficiently converted into the corresponding parameters of the SNN. This also requires to take into consideration the intrinsic differences between the two models, and some adjustments are consequently required to get a correct conversion.
During the training, for each connection among two neurons of the consecutive layers \textit{i} and \textit{i+1},  the weight $w_{i,i+1}$ is learned. Moreover, for each neuron of the layer \textit{i+1}, also the bias $b_{i+1}$ is derived.
In the equivalent SNN model, these parameters need to be translated into an equivalent value for the spiking neural model.
Specifically referring to the Loihi model, the conversion works as follows:
\begin{itemize}[leftmargin=*]
    \item the bias $b_{i+1}$ is associated to the bias current $u_{bias}$ of the neuron $n_{i+1}$. 
    \item $w_{i,i+1}$ is directly set as the weight of the synapse connecting neurons $n_{i}$ and $n_{i+1}$.
\end{itemize}

Besides the learned parameters, each layer of the DNN has to be converted to an equivalent spiking version. This means that each layer will be composed of equivalent spiking neurons that follow the neuron model adopted by the Loihi architecture. 
To apply the DNN-to-SNN conversion, we use the \textit{SNNToolBox} (SNN-TB)~\cite{rueckauer}, an open-source conversion tool that is compatible with Loihi’s Python NxSDK-0.9.5.

The results obtained with the conversion process may not always be optimal, due to several limitations of the NxSDK API and specific constraints of the Loihi neurocores. Therefore, in the following Section~\ref{ANN SNN conversion analysis}, we present a case study for the DNN-to-SNN conversion, specifying a set of general guidelines to follow for achieving a converted SNN that reaches the same accuracy levels as of the corresponding DNN. 

\section{A Comprehensive Analysis on the DNN-to-SNN Conversion Setup}
\label{ANN SNN conversion analysis}
\subsection{Evaluation Metrics for the Conversion Quality}
\label{Preliminar note}
The conversion process requires a series of preliminary considerations for a successful conversion.
First of all, the Loihi architecture uses limited precision synaptic weights, defined within the interval [-256,255]. On the other hand, the trained DNN uses full precision weights. Therefore, a preliminary quantization of the DNN-trained weights is crucial to get a precise converted SNN. 
In this quantization step, the distribution of the input weights has a major role in the outcome of the conversion. That is, the input weights has to be clipped into the Loihi quantized range, therefore a tight weight distribution can be mapped to the quantized interval without relevant errors.
On the other hand, the presence of outliers in the original weights distribution can be the main source of an inprecise conversion. This is due to the fact that very high weights are clipped to fit into the quantized interval, leading to possible inconsistencies between the pre- and post-quantization weight distributions. To decrease strong outliers in the final trained weights, the L2 regularization, applied both on activations and kernels during the training, helps to keep weights into a limited range. 

\begin{figure}[!h]
\vspace*{-5pt}
    \centering
    \includegraphics[scale=0.3]{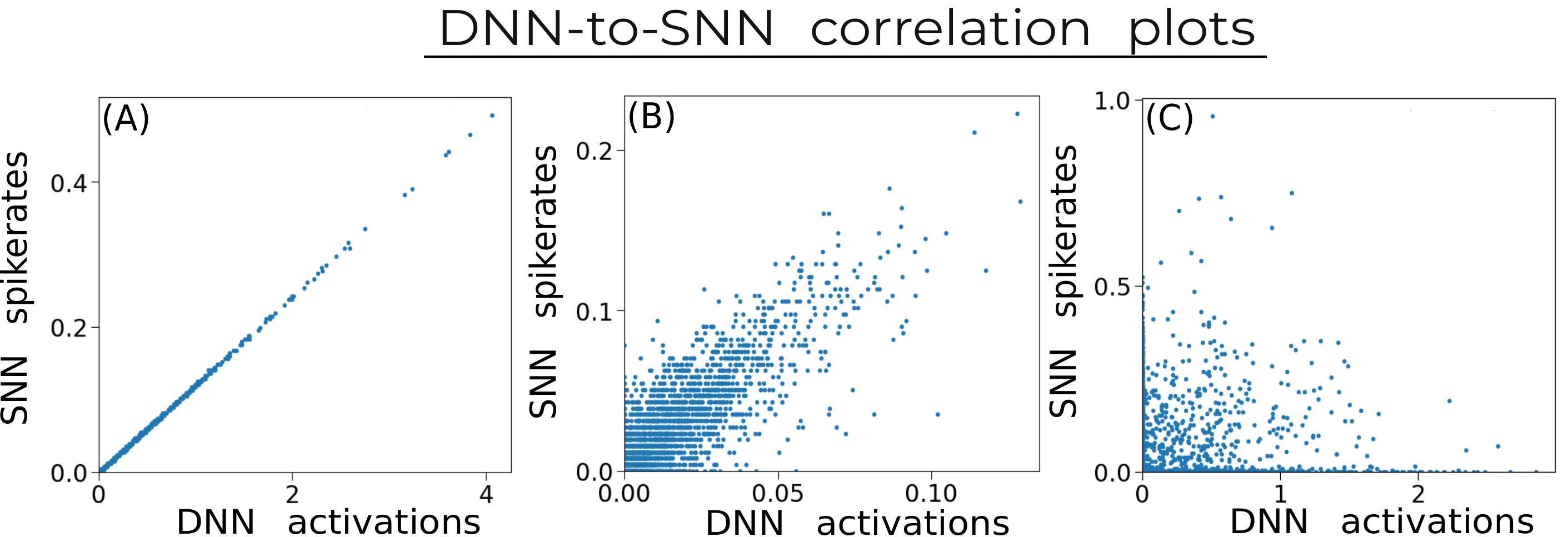}
    \caption{Examples of correlation plots between the DNN activations and their converted SNN spikerates.}
    \label{correlation_plot_example}
    \vspace*{-5pt}
\end{figure}

A good practice to evaluate the quality of the conversion is to look at the \textbf{correlation plots} between the DNN layer activations and the corresponding SNN layer output spikerates. Figure~\ref{correlation_plot_example} shows three typical correlation plots that can be obtained with good and bad conversion processes.

The plot~\ref{correlation_plot_example}(a) is an example of a good correlation plot, where the DNN activations are properly converted into SNN spikerates, being all the points distributed along the main diagonal. 
On the contrary, the plot~\ref{correlation_plot_example}(b) shows a relatively worse conversion, where the DNN activations and the SNN spikerates are still distributed along the diagonal, but the distribution of points is not confined to the desired range. The plot~\ref{correlation_plot_example}(c) is another example of a bad conversion. However, in this case, the activations and the spikerates are totally non-correlated.

\subsection{Tunable Conversion Parameters}
\label{tunable conversion parameters}
Many parameters can be tuned during the DNN-to-SNN conversion process, and a detailed analysis over their effects on the converted SNN is necessary.
These parameters modify the spiking neuron model, the characteristics of the network and the experiment duration.
\begin{itemize}[leftmargin=*]
    \item \textbf{Reset mode}: The reset mode defines the behavior of the neuron after a spike. As previously said, the neuron spikes every time when its membrane potential exceeds the threshold $V_{th}$. After the spike, the membrane potential is reset to a value that depends on the chosen reset mode: 
    \begin{itemize}
        \item \textit{Hard Reset}: The membrane potential is reset to a value equal to 0 after a neuron spikes. This solution is less computationally expensive, but relatively less accurate.
        \item \textit{Soft Reset}: The membrane potential is reset to a value equal to the difference between the highest value reached by the membrane potential and the membrane threshold. This solution is relatively more accurate, but more expensive as well when compared to the hard reset, because the amount of compartments needed to simulate each neuron is doubled.
    \end{itemize} 
    \item \textbf{Desired Threshold to Input Ratio} (\textbf{DThIR}): As described in Section~\ref{sec:background}, the weights of the input DNN model has to be converted to synaptic weights of the SNN. Because of the limited dynamic range of spiking neurons, the output of a spiking neuron may saturate due to an excessively high input, given by some out-of-scale synaptic weights. Hence, it is necessary to normalize the network and set a constant ratio between the incoming neuron inputs and its membrane threshold~\cite{DThIR}. 
    \item \textbf{Experiment duration}: This parameter defines the number of time-steps for which the network receives the same image as an input, i.e., the inference time. A longer duration gives the network more time to output its prediction, but it increases the latency of the system.

\end{itemize}

The development of the DNN architecture has to be realized using the python Keras API~\cite{keras}, which is one of the API supported by Intel NxSDK. Currently, not all the Keras layers are supported by the Loihi's Python NxSDK. The only supported layers are the one in Table~\ref{table_layers}. This limitation has to be taken into consideration during the development of a DNN architecture.

\begin{table}[!h]
\vspace*{-5pt}
    \caption{Layers supported by NxSDK.}
    \vspace*{-3pt}
    \centering
    \begin{tabular}{cccc}
           Dense  &  Flatten & Reshape &  Padding \\ \specialrule{.2em}{.1em}{.1em}
           AvgPooling2D & DepthwiseConv2D & Conv1D &  Conv2D \\ 
    \end{tabular}

    \label{table_layers}
    \vspace*{-12pt}
\end{table}

\subsection{DNN Training}
To study the behavior of the conversion, a small network has been used for evaluating the process.
Such a network, that we will refer to as \textit{cNet}, is a convolutional neural network that contains only convolutional layers and a final dense layer. Its structure is reported in Table~\ref{cNet}.
\begin{table}[!h]
\vspace*{-5pt}
    \caption{\textit{cNet} architecture for the MNIST dataset.}
    \vspace*{-3pt}
    \centering
    \begin{tabular}{c|ccccc}
        
           \textbf{Layer}  &  \textbf{features} & \textbf{Kernel} &  \textbf{stride} & \textbf{Output Shape}  & \textbf{Activation} \\ \hline
           Input          &   1                &                &                  & 28x28x1               & ReLU               \\ 
           Conv2D         &   16               & 4x4            & 2                & 13x13x16              & ReLU               \\ 
           Conv2D         &   32               & 3x3            & 1                & 11x11x32              & ReLU               \\ 
           Conv2D         &   64               & 3x3            & 2                & 5x5x64                & ReLU               \\ 
           Conv2D         &   10               & 4x4            & 1                & 2x2x10                & ReLU               \\ 
           Flatten        &                    &                &                  & 40                    &                \\ 
           Dense          &                    &                &                  & 10                    & SoftMax            \\ 
    \end{tabular}
    \label{cNet}
    \vspace*{-5pt}
\end{table}

To achieve a better conversion process, both activation and weight \textit{L2 Reguralization} are applied on the network layers. In both cases, the value is set to $1\cdot10^{-4}$. The use of regularization during training is preferable for preventing the divergence of the parameter distribution and for avoiding the information loss due to the quantization process of the parameters, as discussed in Section~\ref{Preliminar note}.

The datasets on which the analyses have been performed are the MNIST~\cite{MNIST} and CIFAR10~\cite{CIFAR10}. For each input image, the intensity values are normalized between 0 and 1.
Both networks are developed in Keras, using TensorFlow~\cite{tensorflow} as the backend. The training is performed with the following policies:
\begin{itemize}[leftmargin=*]
    \item \textit{learning rate decay:} initially set to 0.001, it is halved after 15 consecutive epochs without validation accuracy improvements, until it reaches a final value of $5 \cdot 10^{-7}$.
    \item \textit{Adam optimizer}~\cite{Adam}.
    \item \textit{Small data augmentations}, with width and height shifts of 0.1, and 10° rotations.
\end{itemize}

After training, the values of test accuracy achieved by the networks are reported in Table~\ref{accuracyMNIST_CIFAR}.

\begin{table}[h!]
\vspace*{-15pt}
\noindent
\begin{minipage}[t]{.5\linewidth}
\begin{table}[H]
    \caption{Accuracy results of the DNN models.}
    \vspace*{-3pt}
	\centering
	\resizebox{\linewidth}{!}{%
    \begin{tabular}{c|cc}
       \textbf{Nework}  &  \textbf{Dataset} & \textbf{Accuracy}  \\ 	\specialrule{.2em}{.1em}{.1em}
       cNet           &   MNIST   & 98.79\% \\             
       cNet           &   CIFAR10 & 78.92\%     \\         
    \end{tabular}
	}
    \vspace{3mm}
    \label{accuracyMNIST_CIFAR}
\end{table}
\end{minipage}
\hfill
\begin{minipage}[t]{.4\linewidth}
\begin{table}[H]
    \caption{Constraints of the Loihi neurocores.}
    \vspace*{-3pt}
	\centering
	\resizebox{\linewidth}{!}{%
    \begin{tabular}{c|c}
        
           \multicolumn{2}{c}{\textbf{Neurocore constraints}}\\ 	
           \specialrule{.2em}{.1em}{.1em}
           max compartments & 1024 \\
           max fan-in axons        &    4096 \\             
           max fan-out axons    & 4096\\       
    \end{tabular}
	}

    \label{neurocore constraints}
\end{table}
\end{minipage}
\vspace*{-20pt}
\end{table}

\subsection{Conversion Process}
The trained DNN model is then converted into its equivalent spiking model via the SNN-TB tool. The conversion requires four main steps:
\begin{itemize}[leftmargin=*]
    \item \textbf{Parsing}: The toolbox extracts the relevant informations from the original model, discarding layers that are not used in the inference stage (Dropout, BatchNormalization, etc.) and converting the MaxPooling2D layers that may be present into AveragePooling2D, which are supported. The parsed model is the one used as reference for the following conversion.
    \item \textbf{Conversion}: An NxSDK-compatible spiking model is obtained, applying a normalization process that adapts the weights and biases to the limited dynamic range of the spiking neurons, satisfying the selected value of \textit{DThIR}. 
    \item \textbf{Partition}: The conversion process requires to find a valid partition of the neural network on the Loihi chip. Some constraints have to be respected in order to have a valid partition. These constraints, reported in Table~\ref{neurocore constraints}, are related to the synaptic fan-in and fan-out of each neurocore, and the maximum number of neurons that can be mapped onto a single neurocore.
    \item \textbf{Mapping}: The partition is mapped onto the Loihi chip, and the model is now ready to be used in the SNN deployment. 
\end{itemize}

\subsection{Experimental Setup}
\begin{figure}[!h]
    \centering
    \includegraphics[scale=0.8]{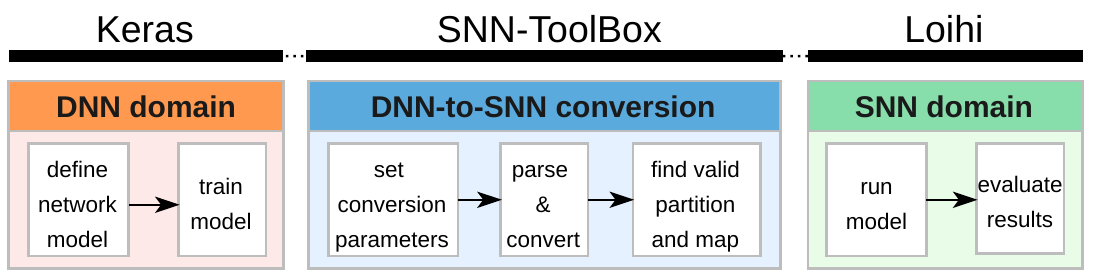}
    \caption{Tool flow of our experimental process.}
    \label{SNN-TB_process}
    \vspace*{-15pt}
\end{figure}

The tool flow is depicted in Figure~\ref{SNN-TB_process}. 
All the experiments are executed on the Intel Neuromorphic Research Cloud (NRC) server, using one of the avaible Loihi partitions. The reported experiments are executed on the Nahuku32 board, which comprises 32 Loihi chips.
As described in section~\ref{tunable conversion parameters}, the three main parameters that have been analyzed for a fine tuning conversion are the \textit{reset mode}, \textit{DThIR}, and \textit{experiment duration}. Different experiments have been done to evaluate the effects of these parameters on the final SNN accuracy.

\subsection{Results Varying the DThIR}
\begin{figure}[!h]
    \centering
    \vspace*{-5pt}
    \includegraphics[scale=0.48]{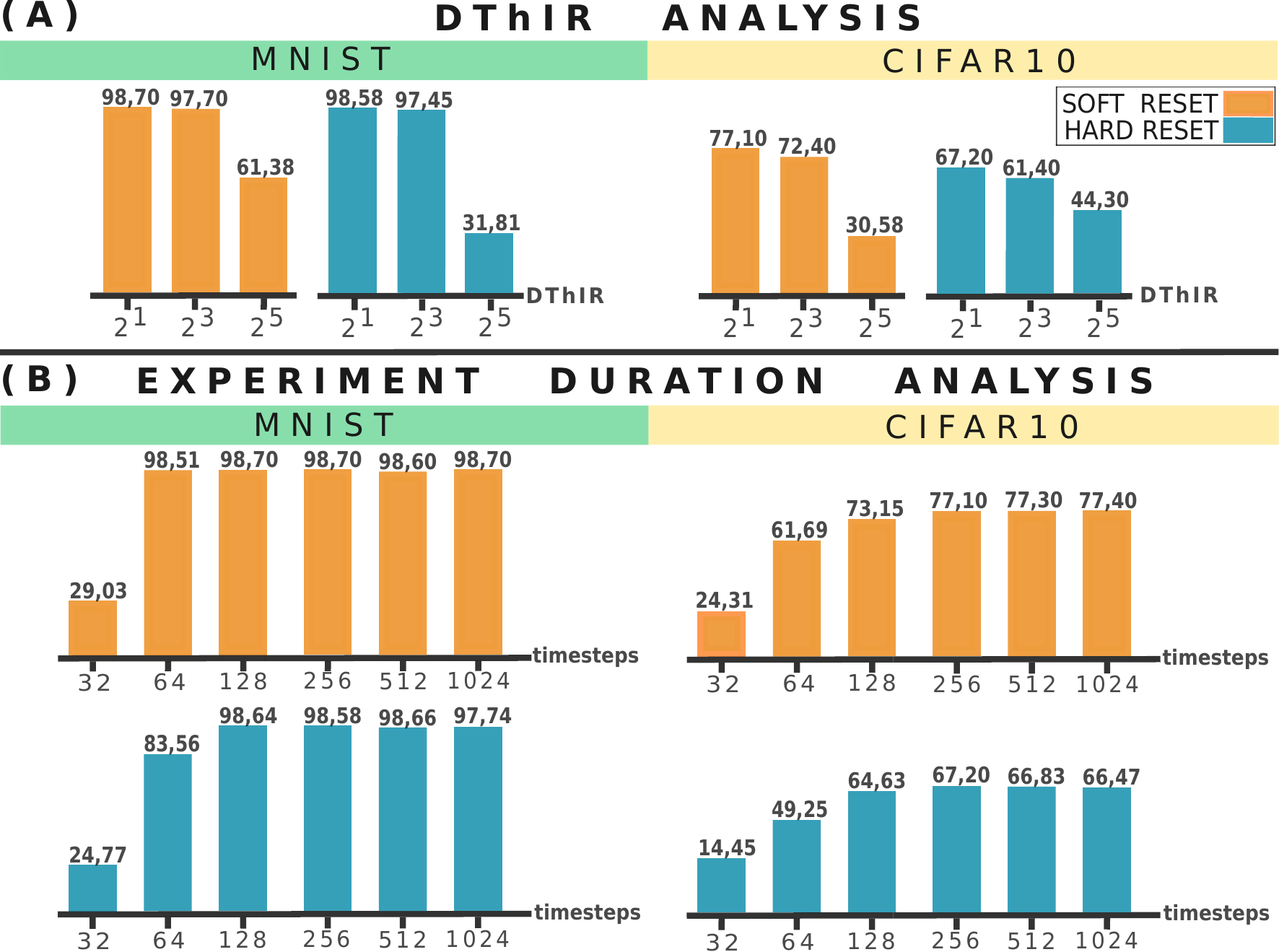}
    \caption{The legend is commong for all the plots. \textit{cNet}, results on MNIST and CIFAR10, varying (a) the DThIR and (b) the experiment duration.}
    \label{results NxSDK MNIST sim duration}
    \vspace*{-5pt}
\end{figure}

In this experiment, we evaluate the conversion results varying the DThIR. The experiment duration is set to 256 time-steps that is a reasonable choice for both the soft and the hard reset, as we will discuss later. 
The tested DThIR levels are $2^1, 2^3$ and $2^5$. Selecting higher levels is usually not a good solution because the membrane potential threshold may get too large. The results are reported in Figure~\ref{results NxSDK MNIST sim duration}(a).

\textbf{Analysis for MNIST}: In both cases of soft reset and hard reset, the SNN accuracy is equal to the DNN accuracy value for DThIR = $2^{1}$ and $2^3$. However, when the parameter is increased to $2^5$, the accuracy drops in both soft and hard reset cases.

\textbf{Analysis for CIFAR10}: Also in this case the highest accuracy is reached for DThIR=$2^1$, for both the hard and the soft reset. However, the accuracy starts reducing when the DThIR is set to $2^3$, and gets to a minimum when the DThIR is increased to $2^5$.

As a consequence of these results, a value of DThIR = $2^1$ is chosen for the following further analysis.

\subsection{Results Varying the Duration and Reset Mode}
This analysis tries to find a good compromise between experiment duration and reset mode. 
By choosing a longer duration, we expect to get more precise results, paying in terms of output latency.
Moreover, the soft reset is expected to provide higher accuracy. 
The results are reported in Figure~\ref{results NxSDK MNIST sim duration}(b).

\textbf{Analysis for MNIST:} Looking at the results achieved on the MNIST dataset, a test accuracy of $98.70\%$ (i.e., only $0.09\%$ lower than the one obtained with the DNN model) is reached in the soft reset case, when the experiment duration is longer than 64 time-steps. On the other hand, it takes at least 128 time-steps for the hard reset case to reach the same level of accuracy. Moreover, the accuracy reached by both the soft and the hard reset remains stable also for longer experiment duration.

\textbf{Analysis for CIFAR10:} The results for the CIFAR10 dataset clearly show that for the hard reset case the DNN accuracy of $78.92\%$ is never reached. The maximum accuracy is $67.20\%$ when the experiment gets longer than 256 time-steps.
On the other hand, the soft reset shows better results than the hard reset, despite not achieving the same results as the corresponding DNN. An accuracy of $77.10\%$ is reached with 256 time-steps, slowly growing to $77.40\%$ with a longer experiment of 1024 time-steps.

For an experiment duration of 256 time-steps, the average time for executing a single inference step of image classification and the Loihi chip usage are reported in Table~\ref{latencies_cNet}. Looking at the number of occupied neurocores, for both the MNIST and CIFAR10 cases, the soft reset makes use of more cores. 

\begin{table}[!h]
\vspace*{-3pt}
    \caption{Accuracy results of the DNN models.}
    \vspace*{-3pt}
    \centering
    \begin{tabular}{c|ccc}
           \textbf{Reset Mode}  &  \textbf{Dataset} & \textbf{Classification time} & \textbf{Neurocores} \\ \hline
           soft           &   MNIST   & 8.312 ms  & 27\\             
           hard          &   MNIST   & 6.464 ms   & 20\\           
           soft           &   CIFAR10 & 21.371 ms & 37\\         
           hard          &   CIFAR10 & 26.159 ms  & 29\\       
    \end{tabular}
    
    \label{latencies_cNet}
    \vspace*{-8pt}
\end{table}

\begin{figure*}[!h]
    \centering
    \includegraphics[scale=0.46 ]{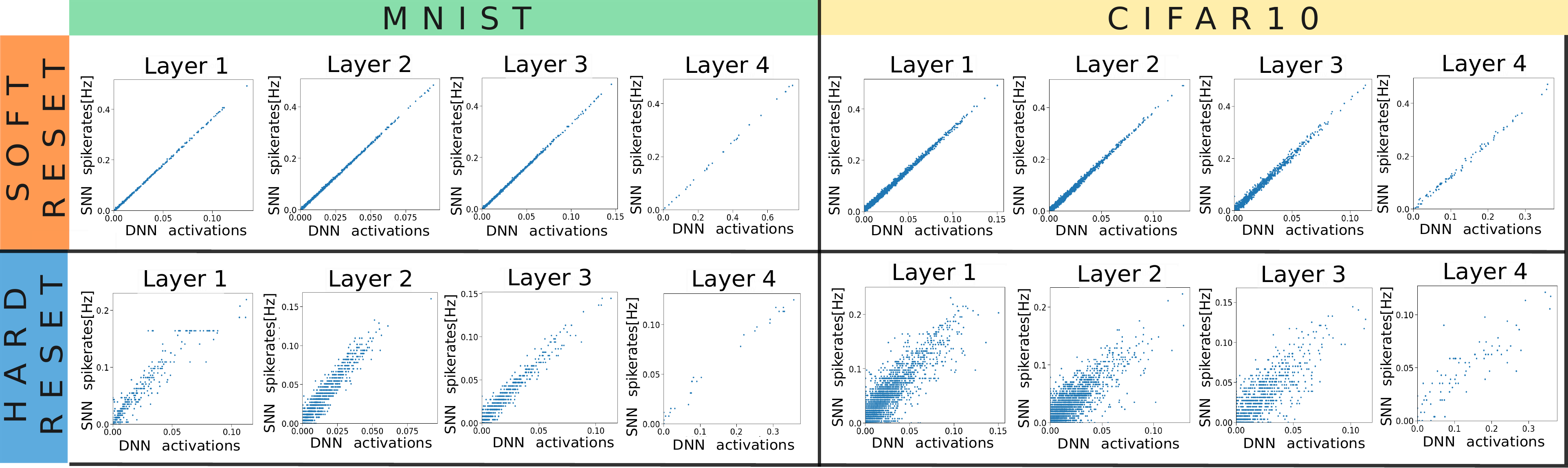}
    \caption{Correlation plots for the first 4 layers of \textit{cNet} after its conversion to the corresponding SNN model. The first column shows the results on the MNIST dataset, whereas the second column presents the results for the CIFAR10 dataset.}
    \label{cNet_correlation_plots}
    \vspace*{-18pt}
\end{figure*}

For better understanding the reason why the soft reset achieves better results than the hard reset conversion, we compare the correlation plots of the converted layers.
Figure~\ref{cNet_correlation_plots} shows the correlation plots of the first 4 layers, both for the soft reset and the hard reset versions, and on both datasets. In each of the 4 presented cases, an experiment duration of 256 time-steps is applied, as well as a DThIR equal to $2^1$.

At a first glance, it is immediately clear that the correlation plots of the soft reset conversion are far more compliant with the expected behavior when compared to the hard reset results, both for the MNIST and CIFAR10 datasets.
Looking at the MNIST - soft reset experiment, the correlation plot of the first layer shows a perfect conglomeration of activations (x axis) vs. spikerates (y axis), along the main diagonal. This means that the conversion of the layer is working as desired, having all the SNN neurons spiking with a rate equivalent to their corresponding DNN activations. The same principle is adopted for the following layers. 

Looking at the MNIST - hard reset experiment, the correlation plots show a relatively worse conversion result.
Starting from the first layer, the points are distributed with a overlapped-staircase behavior. The same happens in the second layer, where it is also present a dilatation of the agglomerate of points along the x-axis. However, both in the $3^{rd}$ and $4^{th}$ layers correlation plots, the points are sufficiently compacted along the diagonal, and in fact the final accuracy achieved by this SNN is similar to the DNN accuracy.

Regarding the CIFAR10 analysis, the soft reset gives good correlation plots, even if the points form a thicker agglomerate w.r.t. the MNIST case. 
On the other hand, the hard reset gives worse results. The correlation between activations and spikerates is relatively less evident, with a general behavior that follows the one of MNIST case, but more emphasized. The analyses reported in these plots justify the 10\% accuracy drop obtained using the hard reset conversion, as seen in Figure~\ref{results NxSDK MNIST sim duration}.

Overall, the results obtained for the CIFAR10 dataset are worse than the ones obtained on the MNIST, both for the soft and the hard reset. This can be addressed to the higher complexity of the CIFAR10, which represents a challenging dataset to work with.

\vspace*{-5pt}

\subsection{Results Discussion}
\label{ANN to SNN Conclusion}

\vspace*{-2pt}

Overall, the use of the \textit{soft reset} mode gives higher accuracy results, because of the lower information loss that occurs during the conversion, as clearly shown by the correlation plots in Figure~\ref{cNet_correlation_plots}. A good choice for the experiment duration seems to be $\geq256$ time-steps. A shorter experiment may lead to an accuracy loss, as shown for the CIFAR10 dataset. On the other hand, using more than 512 time-steps does not lead to a higher level of accuracy, as shown in both the MNIST and CIFAR10 analyses.
Finally, a DThIR value equal to $2^1$ seems to be the best choice to reduce the loss during the conversion. 

Furthermore, the conversion results are also strongly influenced by the DNN architecture, as well as by the DNN training policies.
To have a deeper evaluation of the conversion process, several other DNN models have been trained and converted. These models vary in terms of size, number of layers, and layers characteristics. Not always the conversion process has shown successful results, even applying the soft reset. The problems generally arise when the DNN layers are too wide, making the conversion infeasible, because the neurocore constraints are violated. Therefore, when it comes to build very large SNNs, it is suggested to use depthwise separable convolutional layers that require less core occupation than the traditional ones.
However, a comprehensive analysis concerning these DNN characteristics is not easily practicable, and it is considered beyond the scope of this article.

\vspace*{-5pt}

\section{Pre-processing methods for the DvsGesture dataset}
~\label{dvs_gesture_pre-processing}
The IBM DvsGesture dataset~\cite{DVSgesture} is a fully event-based gesture recognition dataset. Each gesture is recorded with a DVS128 camera, providing a total of 1342 samples divided in 122 trials. In each trial, 1 subject executes the 11 different gestures in sequence. A total of 29 subjects under 3 different light conditions form the whole dataset. Each gesture has an average duration of 6 seconds, and is composed of a collection of all the events (positive and negative) that have been recorded by the DVS camera. A positive (or negative) event is recorded every time a positive (or negative) variation of light is detected.

Event-based data are ideal when used as an input to the SNNs, thanks to their intrinsic asynchronous and spiking behavior. 
However, in the context of our research, we are training a network in the DNN domain, and only at the second stage we convert it into the SNN domain. This forces us to find an alternative representation of the input data, being the DNN not trainable on pure sequences of events.
A valid solution can be to train the DNN with a series of frames obtained by collecting the incoming events.
However, some choices have to be made to achieve a good conversion into frames, that is: 
\vspace{-3pt}
\begin{itemize}[leftmargin=*]
    \item Choose the amount of events to collect into a single frame.
    \item Select the size of the frame and its number of channels.
    \item Set a policy for positive and negative events accumulation.
\end{itemize}
\vspace{-6pt}
\subsection{Events Accumulation}
\vspace{-3pt}
As reported in~\cite{ClosingTheAccuracyGap}, there are two accumulation approaches:
\vspace{-10pt}
\begin{itemize}[leftmargin=*]
    \item \textit{Time-based accumulation:} all events that occur in a fixed time window are accumulated in a single frame.
    \item \textit{Quantitative-based accumulation:} a fixed number of consecutive events are accumulated in a single frame.
\end{itemize}
\vspace{-2pt}

The former solution ensures that the timing information within frames is respected. On the other hand, the latter solution guarantees that each frame will have the same amount of information. However, this may not be a good choice when it comes to gesture recordings. In fact, the number of events generated by a gesture in a fixed time window also depends on the type of the gesture itself. Not all the gestures generate the same amount of events per second. 
Therefore, using a quantitative approach, the number of the generated frames generated depends on the number of events produced by the gesture. Gestures with the same time length may lead to a different amount of frames, having different event rates.

As a consequence, the final dataset will result in an imbalance, having a diverse amount of frames per classes, both in the train and test sets. In order to balance the dataset, one may reduce the amount of frames per gesture to a number that is equal for all classes, but this would come out in a drastic reduction of the used information from the original event-based recordings. Hence, based on these considerations, the time-based accumulation is preferable, because it guarantees a balanced dataset. Therefore, the results relative to the quantitative-based accumulation are not discussed in the following section.
\\
\vspace{-15pt}
\subsection{Time Window Size}
\label{section time window size}
\vspace{-3pt}
The amount of events per seconds varies not only from gesture to gesture, but also between different trials of the same gesture. A mean number of 98 events/ms is estimated by evaluating the original dataset over all the available gestures of all the different trials. This information is a relevant starting point in the choice of the time window size that each frame has to cover. 
In this research, the time windows of 60ms, 150ms, 235ms and 300ms are explored. Choosing a time window of less than 60ms would bring it to an insufficient amount of events collected per frame, thus preventing from having a proper classification. On the other hand, an accumulation time of more than 300ms would lead to a total of less than 3 frames per second, that we consider as the minimum for a real-world application.

A single frame may also have more than one channel, each of them covering a subset of the complete time window. For example, a frame covering a window of 300ms can have 3 channels, where each channel covers a sub-windows of 100ms. This solution allows to get frames in which the temporal information is preserved, since the channels cover consecutive time sections.

Moreover, another solution may be to use overlapped frames, i.e., the time windows covered by two consecutive frames are partially overlapped. For example, using an overlap factor of 2 with frames of 300ms, the frames will cover partially overlapped ranges. The first frame will be $[0ms; 300ms]$ and the following frame will cover the range $[150ms; 450ms]$. \textit{There are several advantages in choosing this solution:}
\begin{itemize}[leftmargin=*]
    \item The number of frames generated from the original dataset is multiplied by the overlap factor, leading to a bigger dataset that guarantees better training results.
    \item The frames can cover different time windows, augmenting the temporal information in the dataset.
    \item The system's throughput is multiplied by the overlap factor. 
\end{itemize}
In our experiments, an overlap factor of 2 has been chosen. Using an overlap factor $n>2$ would lead to generating redundant overlapped frames.
On the other hand, a value $n<2$ would reduce the benefits of having overlapped frames.

\subsection{Events Polarity}
Each event carries the x and y position of the detected event, as well as the polarity of the event that can be either positive or negative.
\begin{itemize}[leftmargin=*]
\item The first possibility is to accumulate the positive and negative events in two different channels of the frame, $c_{+}$ and $c_{-}$. 
Both the channel pixels are initialized at 0, and when a positive event is detected, the pixel (x, y, $c_{+}$) is incremented by 1. On the other hand, a negative event increases the pixel (x, y, $c_{-}$) by 1. Finally, the pixel intensities are normalized in the range $[0; 255]$. Since the accumulation of opposite signed events form a trace of the gesture motion over time, this solution prevents the information loss, because the polarity information becomes relevant when the gestures differ only w.r.t. their sense of rotation.
\item The second solution (as inspired from the work of~\cite{ClosingTheAccuracyGap}) is to accumulate all negative and positive events on the same channel, keeping the polarity information. All the pixels are initialized to a mean value of 128, and are incremented or decremented by 1, depending on the polarity of the event.
\item The third possibility (as inspired from the work of~\cite{ClosingTheAccuracyGap}) is to discard the polarity information and collect all the events in a single channel, by simply incrementing the pixel (x, y) every time either a positive or a negative event occurs. 
\end{itemize}

The above-described three solutions have been tested on the DNN, and based on the accuracy achieved, the following considerations can be made. Overall, the best solution has proved to be the third one, in which the polarity is discarded.
The 2-channel accumulation solution has not shown particular improvements on the final accuracy, when compared to the case in which the polarity is discarded. At the same time, having two channels that separately store the polarity comes with a series of drawbacks, such as, an increase in the size of the dataset as well as in the dimension of the DNN. Moreover, the number of neurocores occupied by the converted SNN is higher than using a single channel, and this also impacts on the latency of the system. For this reason, the 2-channel policy can be discarded.
Considering the 1-channel polarity accumulation, the obtained results have shown an accuracy drop of ($\simeq-4\%$) w.r.t. the discarded polarity case. This solution leads to having frames with generally a high level of pixel intensities, being all initialized to a non-zero value, thereby leading to lower classification results. 
For these reasons, in Table~\ref{table results DVS}, only the results achieved without signed polarity accumulation are reported.

\subsection{Frame Size}
Lastly, the dimension of the frame has to be chosen. The original recordings have a dimension of 128x128. However, such a dimension may be too large when used as an input to our converted SNN, leading to a high number of neurocores required to deploy the SNN on Loihi, as well as increasing the latency of the prediction. Therefore, we resized the image to a dimension of 32x32, by applying a preliminary Average Pooling step. This process is also useful to remove the noisy events from the original recordings, thereby producing a input frames that contain only the relevant gesture information.
Also a 64x64 size has been evaluated, but the accuracy results obtained by the DNN did not show any improvement over the 32x32 size. On the other hand, a size of 16x16 would be too small for achieving a good recognition by the DNN.
\\
Another solution, which has been proposed by~\cite{attentionWindow} for the same dataset, is to collect only the events that are inside a 64x64 attention window, which moves and keeps track of the incoming gestures. Then, the Average Pooling is applied on the 64x64 frame, reducing its size to 32x32.

This solution has been evaluated, but the accuracy results were ($\simeq-5\%$) lower than the one achieved with the whole image frame. The reason for such an accuracy drop may be found in the fact that, by shrinking the input window to the area where the actual gesture takes place, the gesture itself is taken out of its contest. In this way, the DNN cannot distinguish between equivalent gestures executed with opposite arms. 

\subsection{Dataset Structure}
\begin{figure}[!h]
    \centering
    \includegraphics[scale=0.42]{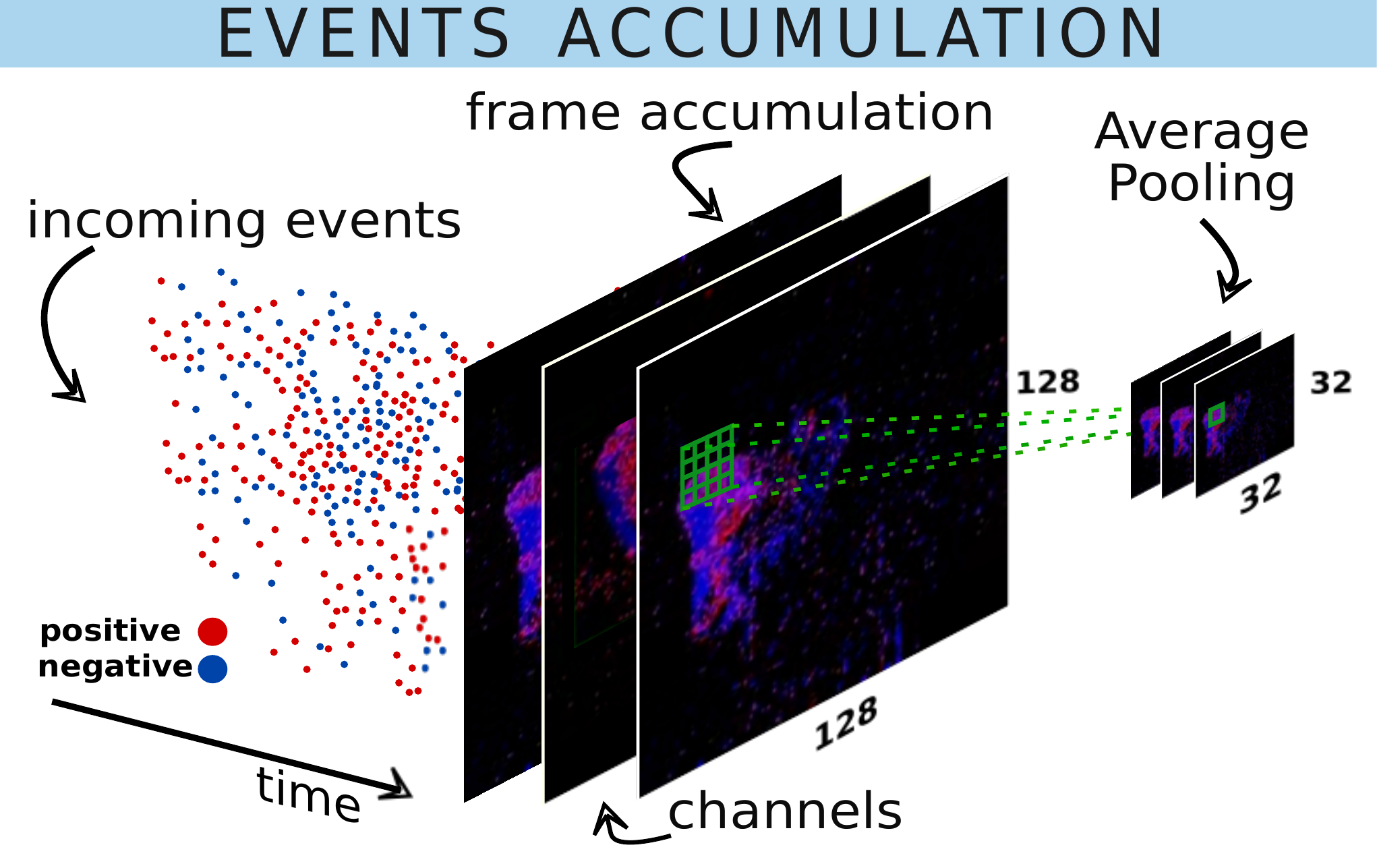}
    \caption{DvsGesture pre-processing: the number of frame channels may depend on the chosen polarity policy or, in a time based accumulation, on the time length of each channel.}
    \label{dvs_preprocessing}
    \vspace*{-13pt}
\end{figure}

In all the above-discussed pre-processing approaches, the frames are associated to their corresponding labels, and accumulated into a \textbf{train set} and a \textbf{test set}. The dimension of the dataset depends on the chosen pre-processing approaches. Less frames are generated with longer time-windows, whereas the amount of frames increases as the time-window covered by each frame gets shorter.
The pre-processing stages are summarized in Figure~\ref{dvs_preprocessing}. 
\vspace*{-5pt}

\section{Accuracy Results}
\label{results DVS}
All the obtained pre-processed datasets have been tested with the \textit{cNet}, the same DNN analyzed in Section~\ref{ANN SNN conversion analysis}, along with the same training parameters for the MNIST and CIFAR10 datasets. This choice has been made to ensure that the possible differences between the DNN and SNN accuracy results depend on the data pre-processing stage, and are not related to the network architecture or the training policies. As explained in Section~\ref{ANN to SNN Conclusion}, if the DNN architecture is modified, the conversion process may suffer and the resulting SNN may not show the expected behavior.

The conversion process has been executed applying the soft reset mode, and an experiment duration of 256 time-steps, with a DThIR=$2^1$, since these are the settings that have shown the best results for both the MNIST and the CIFAR10 analyses. 
Given the analysis provided in Section~\ref{dvs_gesture_pre-processing}, a set of different frame-converted datasets have been realized. In all these datasets, the size of the frame is set to 32x32, and the events polarity is discarded. On the other hand, the converted datasets differ in the frame accumulation time duration, the possible use of the temporal overlapping between frames, and the number of channels per frame.   
Table~\ref{table results DVS} shows the accuracy results for the DNN on the different post-processed datasets.

\begin{table}[!h]
\vspace*{4pt}
    \caption{Pre-processing techniques applied to the original gesture DVS dataset and relative DNN accuracies. For all the datasets, the frame size is equal to 32x32 and the polarity inormation is discarded. All the generated datasets have been tested with the \textit{cNet} DNN.}
    \centering
    \resizebox{\linewidth}{!}{%
    \begin{tabular}{|c?cc|c|c|}
          \hline
           \textbf{Dataset}  &  \textbf{duration(ms)}  & \textbf{overlap}       &  \textbf{channels} &  \textbf{DNN accuracy}\\ \hline
           D1                   &  60 (10 per ch.)           & \xmark      & 6         & 85.23\%            \\ \hline
           D2                  &  60 (20 per ch.)           & \xmark       & 3         & 85.44\%            \\ \hline
           D3                  &  150 (50 per ch. )         & \xmark       & 3         & 87.89\%            \\ \hline
           D4                  &  235 (78 per ch.)          & \xmark       & 3         & 88.63\%            \\ \hline
           D5                  &  300 (100 per ch.)         & \xmark       & 3         & 88.33\%            \\ \hline
           D6                  &  100                       &  \xmark      & 1         & 74.14\%            \\ \hline
           D7                  &  235 (78 per ch.)          & 2            & 3         & 88.87\%            \\ \hline
           D8                  &  300 (100 per ch.)         & 2           & 3         & \textbf{90.46\%}   \\ \hline
    \end{tabular}
    }
    \label{table results DVS}
    \vspace*{-5pt}
\end{table}

Dataset D1 shows that, choosing a time window of only 60ms gives relatively low accuracy results, similar to the case of dataset D2, where the time range covered by each channel is doubled. This can be attributed to a few events accumulated per channel. 

When discussing the datasets D3-5, the time window is progressively incremented, until a maximum duration of 300ms is covered. The results show that a good level of accuracy is reached with a 3-channel frame covering a period of 235ms.

Dataset D6 has been realized to see if using a single channel frame could be a valid solution. In this case, the accuracy drop is evident, and this can be addressed to the fact that the single frame does not contain the temporal information, being all the events accumulated in a single channel.

When discussing the datasets D7 and D8, an overlap factor equal to 2 is introduced. The accuracy increases, reaching a value of \textbf{90.46\%} in dataset D8, which is the best obtained value.
%

The \textit{cNet} DNN model trained on dataset D8 is then converted to its equivalent SNN model representation, and deployed on the Intel Loihi research platform. The converted SNN model reaches a test accuracy of \textbf{89.64\%}, which is only 0.82\% lower than the original DNN model representation. Moreover, the average time for classifying an input frame is \textbf{11.43ms}.
These results have to be compared with the state-of-the-art test accuracies achieved in~\cite{DVSgesture} and in ~\cite{SLAYER}.
The work in~\cite{DVSgesture} reaches a test accuracy of 94.59\% with a 64x64 frame size, whereas the accuracy achieved on a 32x32 frame drops down to 90.78\%. This last value is only 1,14\% higher than the one obtained in this research using frames with the same dimension of 32x32, but it is obtained with a DNN that is much bigger (i.e., 16 convolutional layers with a lot more feature maps per layer) w.r.t. the one used in this work (see Table~\ref{cNet} for our network configuration). However, we did not consider to employ such large and deep networks purposely, in order to maintain a low resource utilization and a low latency for the real-time embedded implementations.

In~\cite{SLAYER}, the test accuracy reached on a smaller portion of the original dataset (1.5 seconds per gesture) is 93.64\%, that is $4\%$ higher w.r.t. the one obtained with our methodology. However, considering the fact that their SNN is designed and trained from scratch (i.e., not a converted one), they have directly used the original event-based dataset, avoiding an inevitable information loss that is related to the pre-processing step.

In terms of latency, with our best solution (D8) the total time needed for a frame classification is $150ms+11.42ms=\textbf{161.42ms}$\footnote{Since the overlap factor is 2, the next frame starts after 150ms, therefore we considered 150ms per frame.}. 
This configuration gives a throughput of 6.24 frames-per-second, which is a feasible solution for a real-time system.

\vspace*{-5pt}

\section{Conclusion}
In this paper, we have proposed an efficient method for deploying gesture recognition through a DVS camera on the Loihi neuromorphic processor. After a careful study of converting a given artificial Deep Neural Network (DNN) to the corresponding Spiking Neural Network (SNN) representation, we devised an efficient pre-processing method for accumulating the events coming from the DvsGesture dataset. As shown by our results, this process enables the training in the DNN domain. Therefore, the well-known training policies and optimizations for DNNs can be employedin this methodology. An efficient conversion of the trained DNN into the SNN domain enables the accurate, energy-efficient and real-time processing on a neuromorphic embedded platform such as the Intel Loihi.

\section*{Acknowledgments}
\begin{small}
\noindent
This work has been partially supported by the Doctoral College Resilient Embedded Systems which is run jointly by TU Wien's Faculty of Informatics and FH-Technikum Wien.
\end{small}

\begin{refsize}
\bibliographystyle{abbrvnat}
\bibliography{main.bib}
\end{refsize}

\end{document}